\documentclass[letterpaper, 10pt, conference]{ieeeconf} 
\IEEEoverridecommandlockouts                          
\usepackage{00_preamble}
\title{\LARGE Geometric-based Line Segment Tracking for HDR Stereo Sequences}

\author{Ruben Gomez-Ojeda, Javier Gonzalez-Jimenez
\thanks{ Machine Perception and Intelligent Robotics (MAPIR) Group, University of Malaga (email: rubengooj@gmail.com). 
\href{http://mapir.isa.uma.es/}{http://mapir.isa.uma.es/}.
}
\thanks{This work has been supported by the Spanish Government (project 
	DPI2017-84827-R and grant BES-2015-071606) and by the Andalucian Government (project TEP2012-530).}
}

\newcommand{\ruben}[1]  {\textcolor{black}{#1}}

\begin{document}

\maketitle
\thispagestyle{empty}
\pagestyle{empty}

\begin{abstract}
In this work, we propose a purely geometrical approach for the robust matching of line segments
for challenging stereo streams with severe illumination changes or High Dynamic Range (HDR) 
environments.
To that purpose, we exploit the univocal nature of the matching problem, 
i.e. every observation must be corresponded with a single feature or not corresponded at all.
We state the problem as a sparse, convex,  $\ell_{1}$-minimization
of the matching vector regularized by the geometric constraints.
This formulation allows for the robust tracking of line segments along sequences 
where traditional appearance-based matching 
techniques tend to fail due to dynamic changes in illumination conditions.
Moreover, the proposed matching algorithm also results in a considerable speed-up of previous 
state of the art techniques making it suitable for real-time applications such as Visual Odometry (VO).
This, of course, comes at expense of a slightly lower number of matches in comparison with appearance-based 
methods, and also limits its application to continuous video sequences, as it is rather constrained 
to small pose increments 
between consecutive frames.
We validate the claimed advantages by first evaluating the matching performance in challenging
video sequences, and then testing the method in a benchmarked point and line based VO algorithm.
\end{abstract}

\newcommand{\lnorm}{$\ell_1$}				
\newcommand{\lnormsq}{$\ell_2$}				

\newcommand {\indfig}{0mm}						
\newcommand {\indtab}{0mm}						
\newcommand {\indsubfig}{0mm}

\newcommand{\wteaser}      {0.21\textwidth}					
\newcommand{\wflst}        {0.3\textwidth}					
\newcommand{\wexperst}     {0.49\textwidth}					
\newcommand{\wexperff}     {0.20\textwidth}					

\newcommand{\setlines}[1]{\mathcal{L}_{#1}}					
\newcommand{\setmatches}[1]{\mathcal{M}_{#1}}				

\newcommand\Tstrut{\rule{0pt}{2.6ex}}         
\newcommand\Bstrut{\rule[-0.9ex]{0pt}{0pt}}   

\newcommand{\wover}      {0.47\textwidth}				
\newcommand{\woptimer}   {0.30\textwidth}				
\newcommand{\wresults}   {0.43\textwidth}				
\newcommand{\wlegend}    {0.30\textwidth}				
\newcommand{\wkitti}     {0.21\textwidth}				
\newcommand{\wtsukuba}   {0.235\textwidth}				
\newcommand{\wtsukubafig}{0.215\textwidth}				

\newcommand{\fig}[1]{Figure \ref{#1}}					
\newcommand{\figs}[2]{Figures \ref{#1} and \ref{#2}}			
\newcommand{\tab}[1]{Table \ref{#1}}					
\newcommand{\secref}[1]{Section {\ref{#1}}}				

\newcommand{\bs}[1]{\boldsymbol{#1}}					
\newcommand{\ssl}[1]{\tensor[^{#1}]}					
\newcommand{\MatrixS}[1]{\bs{#1}}					
\newcommand{\MatrixL}[1]{\textbf{#1}}					
\newcommand{\brackets}[1]{\begin{bmatrix}#1\end{bmatrix}}		

\newcommand{\symmin}[1]{\underset{#1}{\operatorname{min}}}		
\newcommand{\argmin}[1]{\underset{#1}{\operatorname{argmin}}}		
\newcommand{\argmax}[1]{\underset{#1}{\operatorname{argmax}}}		
\newcommand{\der}[2]{\frac{\partial #1}{\partial #2}}			
\newcommand{\derin}[3]{\left.\der{#1}{#2}\right|_{#3}} 			
\newcommand{\prob}[2]{p\left( #1 | #2 \right)}				
\newcommand{\norm}[1]{\left\lVert#1\right\rVert} 			
\newcommand{\fnorm}[1]{\left\lVert#1\right\rVert_{\mathfrak{F}}}	
\newcommand{\skewmat}[1]{ \left[#1\right]_\times }			

\newcommand{\IdMat}{\MatrixL{I}}					
\newcommand{\canonicalvec}[1]{\textbf{e}_{#1}}		
\newcommand{\TRANSPOSE}{^\top}						
\newcommand{\symcov}{\MatrixS{\Sigma}}				
\newcommand{\symRe}{\mathbb{R}}						
\newcommand{\symSSpace}{S}						    
\newcommand{\symPSpace}{\mathbb{P}}					
\newcommand{\symRotSpace}{SO(3)}					
\newcommand{\symRotLie}{\mathfrak{so}(3)}			
\newcommand{\symEucSpace}{SE(3)}					
\newcommand{\symEucLie}{\mathfrak{se}(3)}			
\newcommand{\symrot}{\textbf{R}}					
\newcommand{\symtrans}{\textbf{t}}					

\newcommand{\idL}{L}				
\newcommand{\idR}{R}				
\newcommand{\idF}{k}				
\newcommand{\idFn}{k+1}				

\newcommand{\lIm}{\textbf{l}}			
\newcommand{\normL}{\eta_l}			
\newcommand{\lImFirst}{\lIm_{\idL,\idF}}	
\newcommand{\lImSecond}{\lIm_{\idR,\idF}}	
\newcommand{\lImThird}{\lIm_{\idL,\idFn}}	
\newcommand{\lImFourth}{\lIm_{\idR,\idFn}}	

\newcommand{\match}{\textbf{m}}			

\newcommand{\cam}{C}				
\newcommand{\calib}{\MatrixL{K}}		
\newcommand{\LO}{LO}				
\newcommand{\spoint}{\textbf{p}}		
\newcommand{\epoint}{\textbf{q}}		
\newcommand{\Spoint}{\textbf{P}}		
\newcommand{\Epoint}{\textbf{Q}}		
\newcommand{\spointx}{p_x}		
\newcommand{\spointy}{p_y}		

\newcommand{\separam}{\bs{\xi}}
\newcommand{\separaminc}{\bs{\varepsilon}}
\newcommand{\separamopt}{\separam^*}
\newcommand{\reltrans}{\MatrixL{T}(\separam)}
\newcommand{\reltransopt}{\MatrixL{T}(\separamopt)}

\newcommand{\symover}{\gamma}
\newcommand{\errfun}{\MatrixL{E}}
\newcommand{\weifun}{\MatrixL{W}}
\newcommand{\jacfun}{\MatrixL{J}}

\newcommand{\cross}{\times}	
\newcommand{\lx}{(p_y-q_y)}
\newcommand{\ly}{(q_x-p_x)}
\newcommand{\lbeta}{(p_x^2+p_y^2+q_x^2+q_y^2Id)-2(p_xq_x+p_yq_y}

\section{Introduction}
\label{sec_introduction}
\begin{figure}[!htb]
	\centering
	\subfigure[LSD \cite{von2010lsd} + LBD \cite{zhang2013efficient}]{	
		\includegraphics[width=\wteaser]{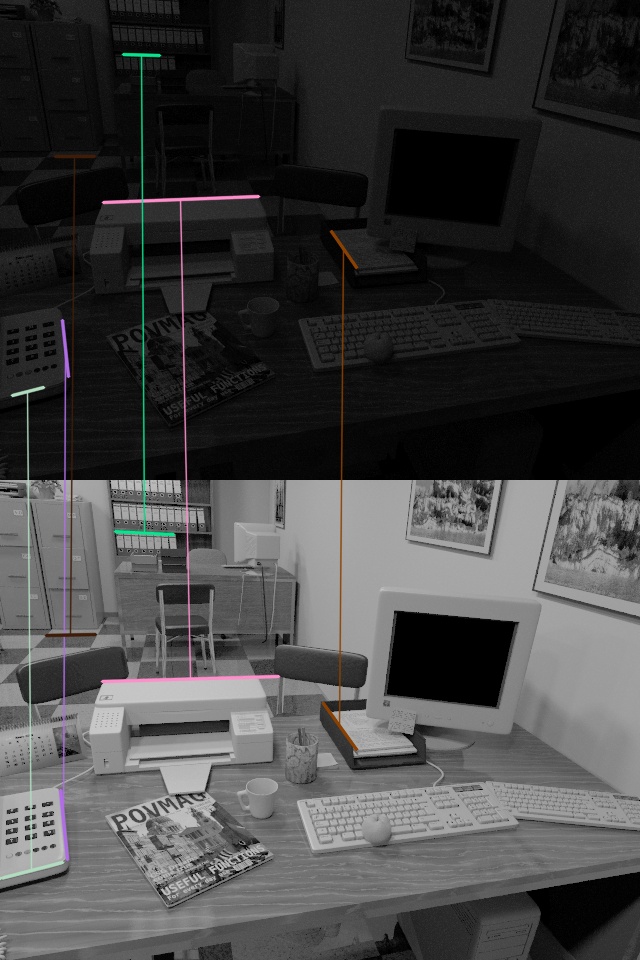}
		\vspace{\indfig}
		\label{lsdlbd}
	}	
	~
	\subfigure[LSD \cite{von2010lsd} + Our proposal]{	
		\includegraphics[width=\wteaser]{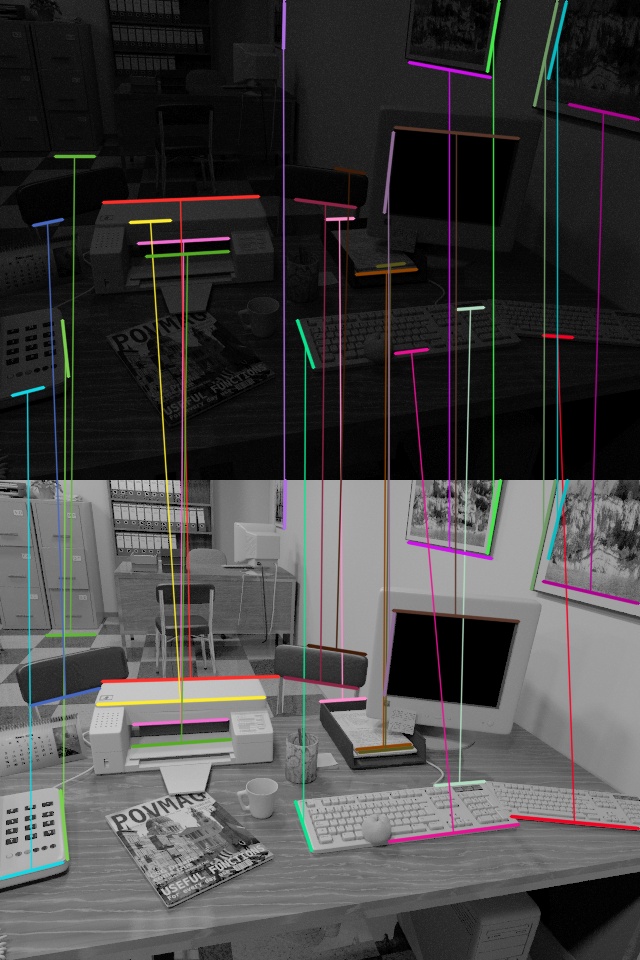}
		\vspace{\indfig}
		\label{lsdfl1}
	}		
	\caption{ 
		Pair of consecutive frames extracted from the sequence \textit{hdr/flicker1} from the dataset in \cite{li2016hdrfusion}
		under challenging illumination changes. Our approach allows for the robust tracking of line segments 
		in this type of environments, where traditional appearance-based matching techniques tend to fail.
		changes in illumination conditions or in HDR scenarios where traditional appearance-based matching 
		techniques tend to fail.
	}
	\vspace{\indfig}
	\label{fig_teaser}
\end{figure}

Although appearance-based tracking has reached a high maturity for \textit{feature-based} motion estimation,
its robustness in real-world scenarios is still an open challenge.
In this work, we are particularly interested in improving the robustness of visual feature tracking in 
sequences including severe illumination changes or High Dynamic Range (HDR) environments (see \fig{fig_teaser}).
%
%
Under these circumstances, traditional descriptors based on local appearance, such as ORB \cite{rublee2011orb} and LBD 
\cite{zhang2013efficient} for points and line segments, respectively, tend to provide many outliers and a low number matches, and hence jeopardizing the performance of the visual tracker.

We claim that line segments can be successfully tracked along video sequences by only considering 
their geometric consistency along consecutive frames, namely, the oriented direction in the image, 
the overlap between them, and the epipolar constraints.
%
%
To achieve robust matches from this reduced segment description we need to introduce some mechanism to deal with
the ambiguity associated to such purely geometrical line matching.

%
For that, we state the problem
as a \textit{sparse}, \textit{convex} \lnorm-minimization of the geometrical constraints from any 
line segment in the first image over all the candidates in the second one, within a \textit{one-to-many} scheme.
This formulation allows for the successful tracking of line segments as it only accepts matches that are
guaranteed to be globally unique.
%
%
In addition, the proposed method results in a considerable speed-up of the tracking process in comparison with
traditional appearance based methods.
For this reason we believe this method can be a suitable choice for motion estimation
algorithms intended to work in challenging environments, even as a recovery stage 
\ruben{
when traditional descriptor-based matching fails or does not provide enough correspondences.
}

\ruben{
To deal with outliers we impose some requiring constraints 
(e.g. small baseline between the two consecutive images) 
}
which slightly reduce the effectiveness of line matching in scenes with repetitive structures and also
the number of tracked features.

In summary, the contributions of this paper are the following:
\begin{itemize}[$\circ$]
\item A novel technique for the tracking of line segments along continuous sequences based on a 
\textit{sparse}, \textit{convex} \lnorm-minimization of geometrical constraints, hence allowing 
for robust matching under severe appearance variations (see \fig{fig_teaser}).
\item A efficient implementation of the proposed method yielding a less computationally demanding 
line-segment tracker which reduces one of the major drawbacks of working with these features.
%
\item Its validation in our previous point and line features stereo visual odometry system 
\cite{gomez2016robust}, resulting in a more robust VO system under difficult illumination conditions, 
and also reducing the computational burden of the algorithm.
%
\end{itemize}
%
%
These contributions are validated with extensive experimentation in several datasets from a wide variety of environments,
where we first compare the accuracy and precision of the proposed tracking technique, and then show 
its performance alongside a VO framework.

\section{Related Work}
\label{sec_related}

%
%
%

Feature-based motion reconstruction techniques, e.g. VO, visual SLAM, or SfM, are typically 
addressed by detecting and tracking several geometrical features (over one or several frames) and 
then minimizing the reprojection error to recover the camera pose.
In this context, several successful approaches have been proposed, such as PTAM \cite{klein2007parallel},
a monocular SLAM algorithm that relies on FAST corners and SSD search over a predicted patch in a
coarse-to-fine scheme for feature tracking.
More recently, ORB-SLAM \cite{mur2017orb} contributed with a very efficient and accurate SLAM system 
based on a very robust local bundle adjustment stage thanks to its fast and continuous tracking of 
keypoints for which they relied on ORB features \cite{rublee2011orb}.
Unfortunately, 
\ruben{
even-though binary descriptors are relatively robust to brightness changes, these techniques 
suffer dramatically when traversing poorly textured scenarios
or severe illumination changes occur (see \fig{fig_teaser}), as the number of tracked features drops.
}

Some works try to overcome the first situation by combining different types of geometric features,
such as edges \cite{Eade2009}, edgelets \cite{forster2017svo}, 
lines \cite{bartoli2005structure}, or planes \cite{ma2016cpa}.
The emergence of specific line-segment detectors and descriptors,
such as LSD \cite{von2010lsd} and LBD \cite{zhang2013efficient}
allowed to perform feature tracking in a similar way as traditionally done with keypoints.
%
%
Among them, in \cite{koletschka2014mevo} authors proposed a stereo VO algorithm relying on image points 
and segments for which they implement a stereo matching algorithm to compute the disparity of several 
points along the line segment, thus dealing with partial occlusions.
In \cite{gomez2016robust} we contribute with a stereo VO system (PLVO) that probabilistically 
combines ORB features and line segments extracted and matched with LSD and LBD by weighting each 
observation with their inverse covariance.
In the SLAM context, the work in \cite{zhang2015building} proposes two different representations:
Plücker line coordinates for the 3D projections, and an orthonormal representation for the motion 
estimation, however, they track features through an optical flow technique, 
thus the performance with fast motion sequences deteriorates.
%
%
Unfortunately, the benefits of employing line segments come at the expense of higher difficulties in dealing with them 
(and they require a high computational burden in both detection and matching stages), and, more importantly,
they still suffer from the same issues as keypoints when working with HDR environments.

A number of methods for dealing with varying illumination conditions have been reported.
%
For example, \cite{engel2017direct} proposed a direct approach to VO, known as DSO, with a joint optimization of 
both the model parameters, the camera motion, and the scene structure. They used the photometric model 
of the camera as well as the affine brightness transfer function to account for the brightness change.
In \cite{zhang16active} authors contributed a robust gradient metric and adjusted the camera setting according to the metric. They designed their exposure control scheme based on the photometric model of the camera and demonstrated improved performance with a state-of-art VO algorithm \cite{forster2014svo}.
Recently, \cite{gomez17learningbased} proposed a deep neural network that embeds images into more informative ones, which are robust to changes in illumination, and showed how the addition of LSTM layers produces more stable 
results by incorporating temporal information to the network.
Although those approaches have proven to be effective to moderate changes in illumination or exposure,
they would still suffer in more challenging scenarios such as the one in \fig{fig_teaser}.


\section{{Geometric-based Line Segment Tracking}}
\label{sec_linetracking}

\subsection{Problem Statement}
The first stage of our segment matching algorithm takes as input a pair of images 
from a stereo video sequence, $I_1$ and $I_2$,
which can be either from the stereo pair or two consecutive ones in the sequence. 
Let us define the sets of line segments 
$\setlines{1} = \{ \textbf{s}_i, \: \textbf{e}_i  \;  |  \:  i \in 1,...,m \}$ and 
$\setlines{2} = \{ \textbf{s}_j, \: \textbf{e}_j  \;  |  \:  j \in 1,...,n \}$ 
in $I_1$ and $I_2$, where we represent the line segment $k$ by their endpoints $\textbf{s}_k$ and $\textbf{e}_k$ in
homogeneous coordinates. We also employ the vector of the line 
\begin{equation}
\vec{\textbf{l}}_k = \frac{\textbf{s}_k - \textbf{e}_k}{\norm{\textbf{s}_k - \textbf{e}_k}_2}
\end{equation}
estimated from the segment endpoints to compare the geometric features of each of them. 

\begin{figure}[!t]
	\centering
	\subfigure[LSD \cite{von2010lsd} + LBD \cite{zhang2013efficient}]{	
		\includegraphics[width=\wexperst]{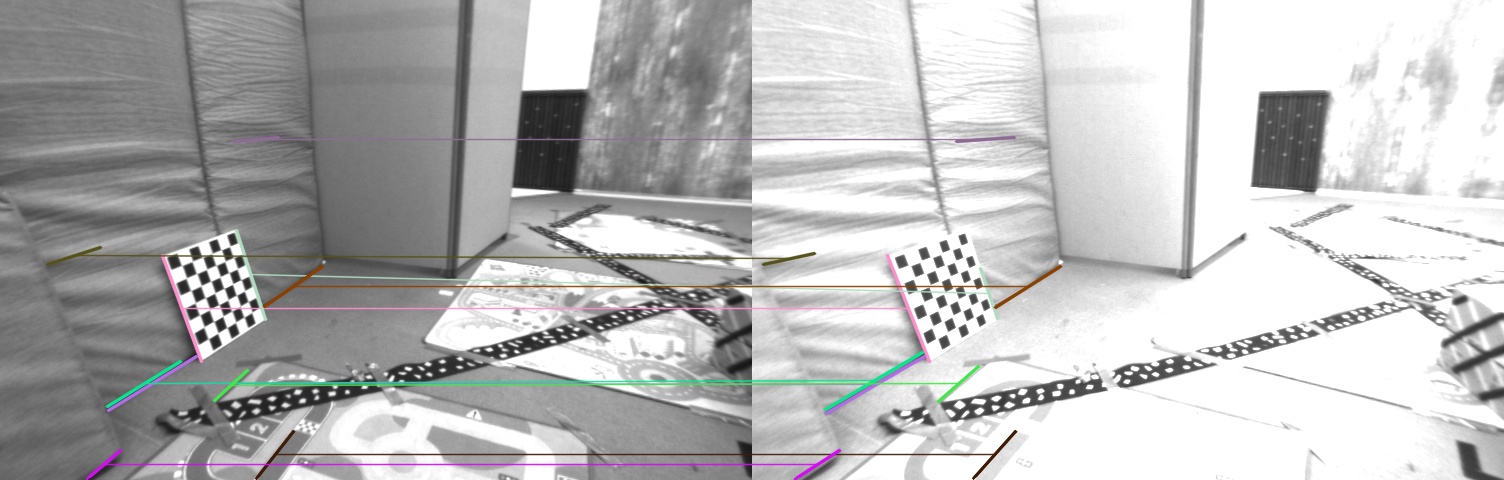}
		\vspace{\indfig}
		\label{euroc_lbd_st}
	}
	\\	
	\subfigure[LSD \cite{von2010lsd} + Our proposal]{	
		\includegraphics[width=\wexperst]{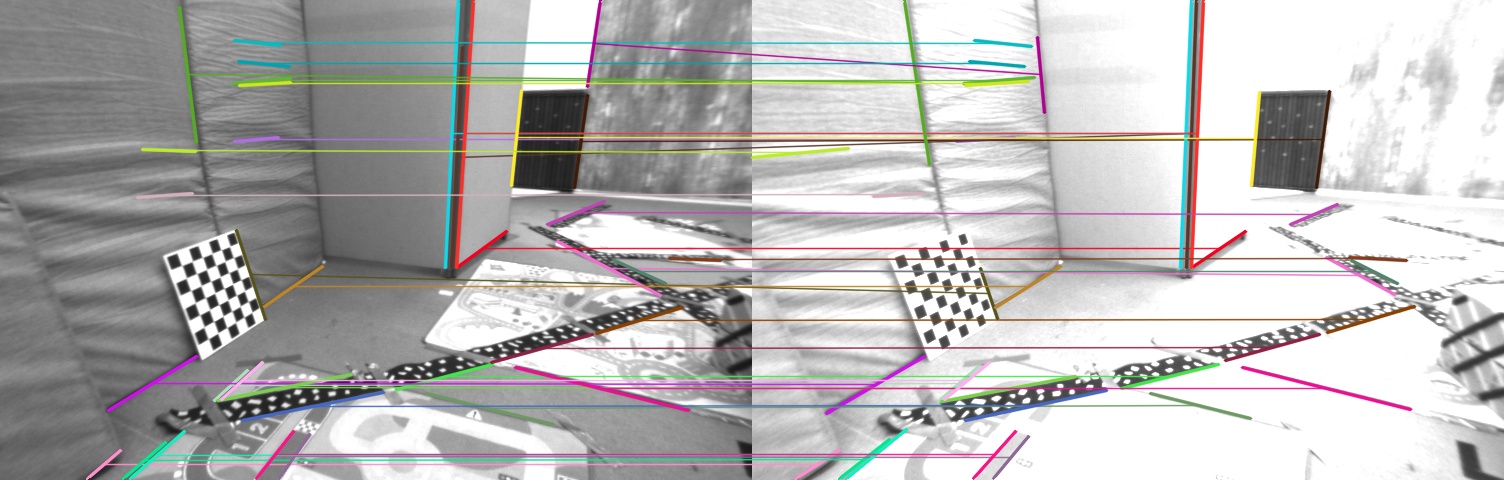}
		\vspace{\indfig}
		\label{euroc_fl1_st}
	}
	\caption{ 
	Stereo correspondences between two different images from the EuRoC dataset. 
	Our matching algorithm is capable of finding matches that does not
	necessarily have similar appearance.
	}
	\vspace{\indfig}
	\label{fig_stereo}
\end{figure}

Then, given $\setlines{1}$ and $\setlines{2}$, our aim is to find the subset of corresponding line segments 
between the two input images (see \fig{fig_stereo}), defined as 
$\setmatches{12}= \{ (l_i,l_j) \;  |  \: l_i \in \setlines{1} \: \wedge \: l_j \in \setlines{2}   \}$.
%
For $l_i$ and $l_j$ to be a positive match,
they must be parallel, have a sufficient ovelap and be compliant with the epipolar geometry of the two views. 
In order to impose the lines to be parallel, we consider the angle formed by the two line segments
in the image plane, $\theta_{ij}$:
%
\begin{equation}
\label{eq_angleconstraint}
\theta_{ij} = atan( \; ||{\vec{ \textbf{l}}_i \times \vec{\textbf{l}}_j }|| \; / \; \vec{ \textbf{l}}_i \cdot \vec{\textbf{l}}_j \; ).
\end{equation}
%

\begin{figure}[]
\centering
\includegraphics[width=\wflst]{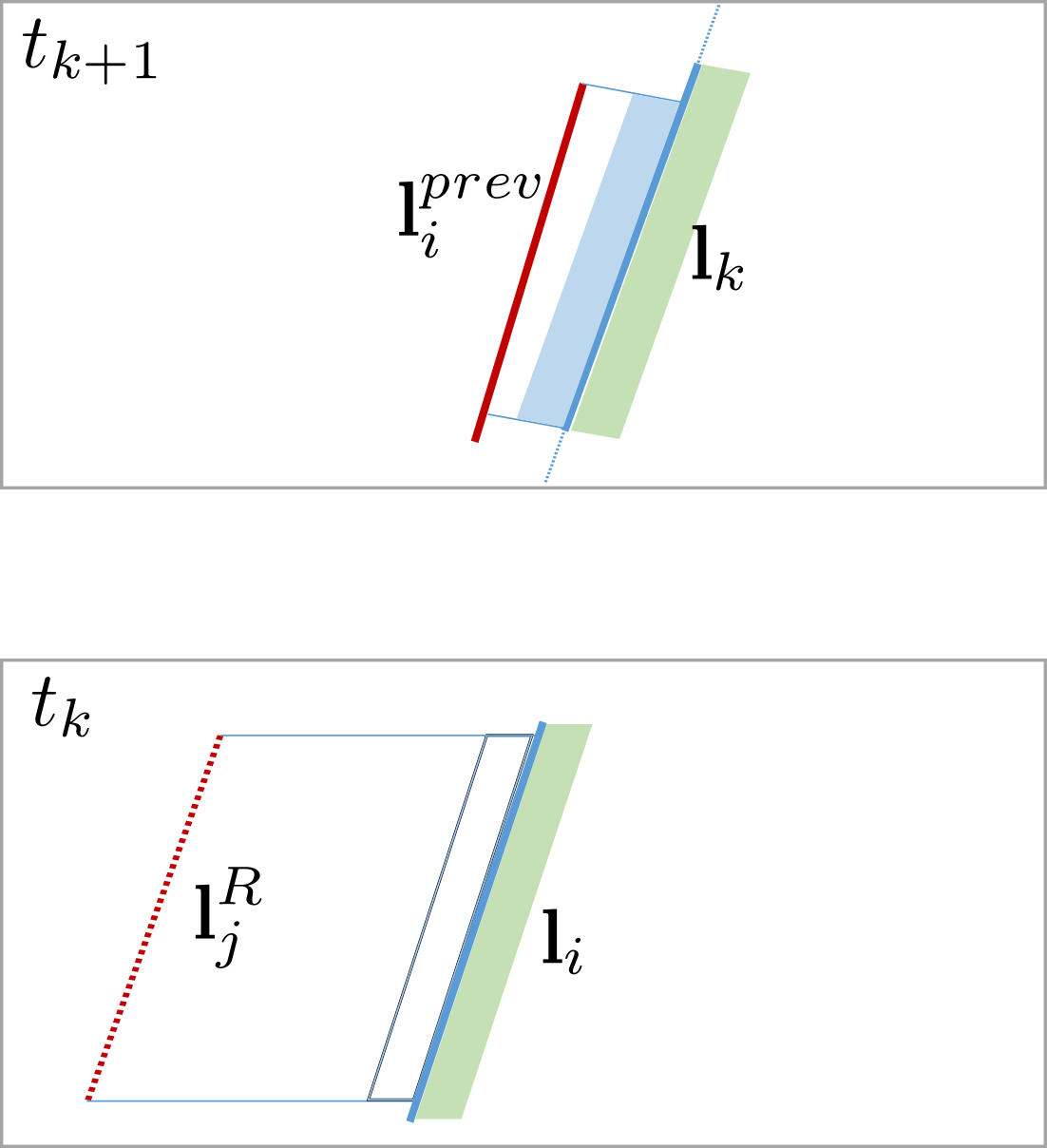}
\caption{
	\ruben{
Scheme of the line segment overlap for both the \textit{stereo} and \textit{frame-to-frame} cases.
In the bottom image $t_k$ we plot the stereo overlap between the reference line in the left image $\textbf{l}_i$
and a match candidate in the right one  $\textbf{l}^R_j$, which is the ratio of the lengths of the shadowed areas (in blue the overlap and in green the line's length).
Similarly, the above image $t_{k+1}$ depicts the overlap between the reference line in the second image $\textbf{l}_k$
and the projected line in the second frame $\textbf{l}^{prev}_i$.
}
}
\vspace{\indfig}
\label{fig_linetracking}
\end{figure}

The above-mentioned expressions, however, might lead to inconsistent results as any line in the image could
satisfy Equation \eqref{eq_angleconstraint} without being related to the
query one.
Therefore, we deal with this phenomena by also defining the \textit{overlap} of two line segments
$\rho_{ij} \in [0,1]$ as the ratio between their common parts, as depicted in \fig{fig_linetracking}, where 
$\rho_{ij}$ equals 0 and 1 when there is none or full overlapping between the line segments, respectively.
In addition, we also define the ratio between the line lengths as:
\begin{equation}
\mu_{ij}  =  \frac{max( L_i, L_j )}{min( L_i, L_j )}
\end{equation}
where $L_k = \norm{\textbf{s}_k - \textbf{e}_k}_2$ stands for the length of the $k$-th line,
which discards any likely pair of segments whose lengths are not similar enough 
(if they are of similar length the value of $\mu_{ij}$ is close to one, and bigger than one otherwise).

Finally, we also consider epipolar geometry as a possible constraint for the two different cases of study.
In the first case, \textit{stereo} matching, we define the angle formed by the middle point flow vector,
$\textbf{x}_{ij} = \textbf{m}_i - \textbf{m}_j $ where the middle point is defined as $\textbf{m}_k = (\textbf{s}_k+\textbf{e}_k)/2$, as:
\begin{equation}
\label{eq_epipstereo}
\theta^{st}_{ij} = asin( \norm{\textbf{x}_{ij} \times \bs\eta_1} / \norm{\textbf{x}_{ij}} )
\end{equation}
where $\bs\eta_1$ stands for the director vector of the $X$ direction.
In contrast, in the \textit{frame-to-frame} case, we assume that images are separated by a small motion
and therefore we define the angle formed by $\textbf{x}_{ij}$ and the $Y$ direction (whose unit vector is given by $\bs\eta_2$), namely:
\begin{equation}
\label{eq_epipf2f}
\theta^{ff}_{ij} = asin( \norm{\textbf{x}_{ij} \times \bs\eta_2} / \norm{\textbf{x}_{ij}} ).
\end{equation}
%

\subsection{Sparse \lnorm-Minimization for Line Segment Tracking}

In this paper, we formulate line-segment tracking as a sparse minimization problem solely based on the previously
introduced geometric constraints.
Although this representation has been already employed in computer vision for 
noise reduction \cite{elad2010role}, face recognition \cite{cheng2010learning}, 
and loop closure detection \cite{latif2017sparse} (among others),
to the best of our knowledge this is the first time it is employed for the geometric 
tracking of line segment features.
For that, we also take advantage of the \textit{1-sparse} nature of the tracking problem, i.e. 
a single line $l_i$ from the first image should only have at most one match candidate 
from the $\setlines{2}$ set.
It must be noticed that, in the case of detecting divided lines, it is possible for
more than one line to match the query one, however, this case is even more likely to 
occur with appearance based methods, as any locally similar line in the image can be a candidate.

Let us define the $n$-dimensional \textit{matching} vector $\bs\omega_{i}$ of the line $l_i \in \setlines{1}$ as:
\begin{equation}
\bs\omega_{i} = \big[ \omega_{i0} \: ... \: \omega_{ij} \: ... \: \omega_{in}	\big] \TRANSPOSE
\end{equation}
where $\omega_{ij}$ equals one if $l_i$ and $l_j$ are positive matches and zero otherwise,
and $n$ stands for the number of line segments in $\setlines{2}$.
Moreover, we define the line segment \textit{error} vectors $\bs\beta_{ij}$
and the objective $b$ for both the \textit{stereo} and \textit{frame-to-frame} cases as:
\begin{equation}
\bs{\beta}_{ij} = 
\left[ 
\begin{array}{c}
\theta_{ij}			\\
\theta_{ij}^{epip}	\\
\rho_{ij}			\\
\mu_{ij}	
\end{array}
\right]
\; , \;\;\;
\textbf{b} = 
\left[ 
\begin{array}{c}
0   \\
0   \\
1   \\ 
1
\end{array}
\right]
\end{equation}
for the line segments $l_i \in \setlines{1}$ and $l_j \in \setlines{2}$,
where \textit{epip} refers to the epipolar constraints defined in Equations \eqref{eq_epipstereo} and
\eqref{eq_epipf2f} for the two cases of study.
%

Now, by concatenating all line segment error vectors we form the $4\times n$ matrix $\textbf{A}_i$:
\begin{equation}
\textbf{A}_i = \big[ \bs\beta_{i0}, \; ... \; \bs\beta_{ij}, \; ... \; \bs\beta_{in}	\big].
\end{equation}
that must satisfy the linear constraint 
$\textbf{A}_i \bs\omega_{i}=\textbf{b}$
if the sum over all the components from the matching vector $\bs\omega_{i}$  is one 
(which is our hypothesis).
While \lnormsq-norm is usually employed to solve the previous problem with the typical least-squares
formulation, it is worth noticing that it leads to a dense representation of the optimal $\bs\omega_{i}^{*}$, 
which contradicts the 1-sparse nature of our solution.

In contrast, we can formulate the problem of finding $l_j \in \setlines{2}$ that properly matches $l_i \in \setlines{1}$
as a \textit{convex}, \textit{sparse}, \textit{constrained} \lnorm-minimization as follows:
\begin{equation}
\label{eq_constrained}
\symmin{ \bs\omega_{i} } \; 
\norm{\bs\omega_{i}}_1 \; 
\textnormal{subject to} \; 
\norm{ \textbf{A}_i \bs\omega_{i} - \textbf{b}   }_2 \le \epsilon
\end{equation} 
where the constraint corresponds to the above-mentioned geometrical conditions,
and $\epsilon > 0 $ is the maximum tolerance for the constraint error.
Moreover, the problem in Equation \eqref{eq_constrained} can be also solved with the homotopy approach
\cite{asif2008primal} in the following \textit{unconstrained} manner:
\begin{equation}
\label{eq_unconstrained}
\symmin{ \bs\omega_{i} } \; \;
\lambda \norm{\bs\omega_{i}}_1 \; 
+ \; \frac{1}{2}
\norm{   \textbf{A}_i \bs\omega_{i} - \textbf{b}   }_2
\end{equation} 
with $\lambda$ a weighting parameter empirically set to 0.1, 
 resulting in a very effective and fast solver \cite{donoho2008fast}.   

Then, we efficiently solve the problem in Equation \eqref{eq_unconstrained} for each $l_i \in \setlines{1}$ 
obtaining the sparse 
vector $\bs\omega_{i}$, 
which after being normalized indicates whether the line segment $l_i$ has
a positive match (in the maximum entry $j$ of $\bs\omega_{i}$).
Finally, we guarantee that line segments are uniquely corresponded
by only considering the candidate with minimum error, defined as $\norm{\bs{\beta}_{ij}}$, 
if the error for the second best match is at least 2 times bigger than the best one.
For further details on the mathematics of this Section, please refer to \cite{asif2008primal}.

\subsection{Dealing with Outliers}

When dealing with repetitive structures a number of outliers can appear.
To deal with this problem in the \textit{stereo} case, we implement a filter based on the epipolar constraint,
for which we first estimate robustly the normal distribution formed by the angles with the horizontal
direction.
Then, we discard the matches whose angle with the horizontal direction lies above 2 times the standard
deviation of the distribution formed by all matches.

In the \textit{frame-to-frame} case, as the camera pose
is not known yet, we cannot directly apply epipolar geometry.
However, we approximate an epipolar filter, based on the assumption that input images 
belong to consecutive frames from a sequence, and therefore they are separated by small motions.
For that, we discard the matches whose angle with the vertical direction (this is the epipolar constraint in the
case of null motion)  lies above 2 times the standard deviation of the distribution formed by all matches
\ruben{
as they are less likely to fullfil the motion constraints.
}


\section{Point-Segment Visual Odometry Overview}
\label{sec_overview}

In this section we briefly describe the PLVO stereo visual odometry system \cite{gomez2016robust}
where the proposed matching algorithm has been integrated for line segment tracking. 
PLVO combines probabilistically both point and line segment features and its C++ implementation is 
available publicly \url{https://github.com/rubengooj/StVO-PL}.
%

\subsubsection{Point Features}
In PLVO points are detected and described with ORB \cite{rublee2011orb}
(consisting of a FAST keypoint detector and a BRIEF descriptor) due to its 
efficiency and good performance.
In order to reduce the number of outliers, we only consider the measurements that are mutual best matches,
and also check that the two best matches are significantly separated in the description space
by only accepting matches whose distance between the two closest correspondences
is above the double of the distance to the best match.

\subsubsection{Line Segment Features}
In our previous work \cite{gomez2016robust} we detect line segments with the 
Line Segment Detector (LSD) \cite{von2010lsd} and also employ the
Line Band Descriptor (LBD) \cite{zhang2013efficient} for the stereo and frame-to-frame matching.
Although this method provides a high precision and repeatability, it still presents very
high computational requirements, simple detection and matching requires more than 30ms with $752\times480$,
thus its use limits their application in real-time.
In order to reduce the computational burden of the Stereo VO system, in this work we have also employed
the Fast Line Detector (FLD) \cite{lee2014outdoor}, which is based on connecting collinear Canny edges \cite{canny1986computational}.
This detector works faster than LSD at expense of a poorer performance in detecting meaningful lines,
i.e. lines with strong local support along all the lines, since, unlike LSD, detection is only based on image edges.

\subsubsection{Motion Estimation}
After obtaining a set of point and line correspondences, we then recover the camera motion
with iterative Gauss-Newton minimization of the projection errors in each case (in the case
of line segments we employ the distance from the projected endpoint, to the line in the next frame).
To mitigate the undesirable effect of outliers and noisy measurements, we perform a two
steps minimization for which we weight the observations with a Pseudo-Huber loss function,
and then we remove the outliers and refine the solution.


\section{{Experimental Validation}}
\label{sec_evaluation}
We evaluate the performance and robustness of our proposal in several public datasets 
for two different line segment detectors,
LSD \cite{von2010lsd} and FLD \cite{lee2014outdoor}, when employing two different matching strategies:
our proposal, and traditional appearance based tracking with LBD \cite{zhang2013efficient}.
All the experiments have run on an Intel Core i7-3770 CPU @ 3.40 GHz and 8GB RAM without GPU 
parallelization. 
In our experiments we have employed a fixed number of detected lines set to 100, and 600 ORB \cite{rublee2011orb}
features for the case of points and line based VO.
%
%
%
\begin{table*}[!htb]
\centering
\caption[]{Tracking performance  of our proposal and traditional line segment feature matching 
(number of matches - inliers).}
\label{tab_performance}
\begin{tabular}{lc|cccc|cccc}
Dataset & Resolution & 
\multicolumn{2}{c}  { LSD + LBD } 	& 
\multicolumn{2}{c|} { LSD + L1  }  	&
\multicolumn{2}{c}  { FLD + LBD } 	& 
\multicolumn{2}{c}  { FLD + L1  }  	\\
\hline 
\hline
hdr/bear 		& $640\times480$ 	  	  & 72 & 80 \% & 39 & 94 \% 	& 65 & 73 \% & 45 & 86 \% \Tstrut\\
hdr/desk 		& $640\times480$ 	  	  & 70 & 84 \% & 40 & 93 \% 	& 60 & 83 \% & 41 & 88 \% \\
hdr/floor1 		& $640\times480$ 	  	  & 27 & 77 \% & 22 & 85 \% 	& 29 & 77 \% & 33 & 85 \% \\
hdr/floor2 		& $640\times480$ 	  	  & 71 & 80 \% & 42 & 93 \% 	& 61 & 80 \% & 41 & 87 \% \\
hdr/sofa 		& $640\times480$ 	  	  & 58 & 81 \% & 41 & 90 \% 	& 49 & 81 \% & 45 & 84 \% \\
hdr/whiteboard 	& $640\times480$ 	  	  & 41 & 76 \% & 35 & 95 \% 	& 42 & 75 \% & 35 & 90 \% \\
hdr/flicker1 	& $640\times480$ 	  	  & 42 & 84 \% & 45 & 98 \% 	& 44 & 80 \% & 40 & 95 \% \\
hdr/flicker2 	& $640\times480$ 	     & 86 & 89 \% & 45 & 98 \% 	& 74 & 85 \% & 45 & 97 \% \\
\hline
dnn/1-light			& $752\times480$ 	  & 16 & 94 \% & 15 & 100 \% 	& 19 & 90 \% & 18 & 100 \% \Tstrut\\
dnn/2-lights		& $752\times480$ 	  & 29 & 94 \% & 18 & 100 \% 	& 38 & 92 \% & 24 & 97  \% \\
dnn/3-lights		& $752\times480$ 	  & 49 & 90 \% & 35 & 100 \% 	& 57 & 91 \% & 35 & 97  \% \\
dnn/change-light	& $752\times480$ 	  & 19 & 85 \% & 14 & 100 \% 	& 24 & 95 \% & 18 & 100 \% \\
dnn/hdr1			& $752\times480$ 	  & 55 & 90 \% & 35 & 100 \% 	& 51 & 88 \% & 35 & 95  \% \\
dnn/hdr2			& $752\times480$ 	  & 50 & 90 \% & 32 & 97 \% 	& 52 & 93 \% & 39 & 95  \% \\
dnn/overexp			& $752\times480$ 	  & 84 & 94 \% & 55 & 98 \% 	& 75 & 92 \% & 47 & 96  \% \\
dnn/overexp-change-light & $752\times480$ 
									 	  & 82 & 92 \% & 50 & 98 \% 	& 72 & 90 \% & 43 & 96 \% \\
dnn/low-texture		& $752\times480$ 	  & 53 & 88 \% & 35 & 100 \% 	& 52 & 90 \% & 35 & 97 \% \\
dnn/low-texture-rot	& $752\times480$ 	  & 43 & 90 \% & 30 & 100 \% 	& 40 & 90 \% & 30 & 95 \% \\
\hline
tsukuba 						& $640\times480$ 	  & 43 & 86 \% & 34 & 84 \% & 36 & 75 \% & 21 & 86 \% \Tstrut\\
tsukuba/fluor(L)-daylight(R)	& $640\times480$ 	  & 5  & 40 \% & 15 & 80 \% &  5 & 40 \% & 12 & 80 \% \\
tsukuba/fluor(L)-flashlight(R)	& $640\times480$ 	  & 2  & 33 \% &  5 & 60 \% &  2 & 50 \% &  4 & 50 \% \\
tsukuba/fluor(L)-lamps(R)		& $640\times480$ 	  & 1  &  0 \% &  5 & 60 \% &  1 &  0 \% &  3 & 67 \% \\
%
\end{tabular}
\vspace{\indtab}
\end{table*}

\subsection{Tracking Performance}
First, we compare the line segment tracking performance of our proposal against traditional
feature matching approaches. 
For that, we took several sequences (at different speeds) and 
classified each match as an inlier if the correspondent line segment projection error is less
than one pixel when employing the groundtruth transformation.
In order to compare the algorithms under dynamic illumination changes, 
we have employed two specific datasets: one extracted from \cite{li2016hdrfusion} (\textit{hdr})
taken with an RGB-D sensor under HDR situations, and another one from our previous work
\cite{gomez2017learning} (\textit{dnn}) containing a number of difficult dynamic illumination conditions.
In addition, we also have employed the Tsukuba Stereo Dataset \cite{peris2012towards} , a synthetic dataset
rendered under 4 different illuminations, i.e. \textit{fluorescent}, \textit{lamps}, \textit{flashlight}, and \textit{daylight}.
For a more challenging set of experiments, we have also employed to use all combinations (taking \textit{fluorescent}
as reference)
of the rendered sequences, by setting the left one to the reference and the right one to all different possibilities.
It is worth noticing that illumination changes from the considered datasets are produced punctually, and after that,
the scene illumination usually keeps constant until the next change.
This benefits to descriptor-based techniques when evaluating the tracking performance during the whole sequence, 
for which we also recommend to watch the attached video for visual evaluation under such circumstances.

\tab{tab_performance} shows the tracking accuracy and the number of features tracked, for all the sequences from
each considered dataset.
First, we observe a slightly inferior performance of FLD \cite{lee2014outdoor} in comparison against LSD \cite{von2010lsd},
due to its lower repeatability in contrast with its superior computational performance (see \tab{tab_perform}).
In general, we observe that our matching method decreases the number of features, due to the very requiring
assumptions of our matching technique, however, it provides a higher ratio of inliers thanks to the extra stage 
explained in \secref{sec_linetracking}.

As for the Tsukuba dataset, we observe that the number of features successfully tracked dramatically decreases 
as the response of the detectors is not capable of producing a compatible set of lines from the same images.
However, we observe that our method technique is capable of recovering more matches, specially in the less challenging 
case (\textit{fluorescent} and \textit{daylight}), that can be employed along different sensing to extract more information
from the environment in such difficult situations.

\subsection{Robustness Evaluation in Stereo Visual Odometry}
In this set of experiments, we test the performance of the compared algorithms in 
the EuRoC \cite{burri2016euroc} dataset. 
In order to simulate changes in exposure time or illumination within the EuRoC dataset \cite{burri2016euroc} 
(we will refer to simulated sequences with an asterisk)
we change the gain and bias of the image with two uniform distribution, 
i.e. $\alpha = \mathcal{U}(0.5,2.5)$ and $\beta = \mathcal{U}(0,20)$ pixels every 30 seconds.
For that comparison, we not only focus in the accuracy of the estimated trajectories, but also in the
robustness of the algorithms under different environment conditions (we mark a dash 
those experiments where the algorithm lose the track).
We compare the accuracy of trajectories obtained with our 
previous stereo VO system, PLVO \cite{gomez2016robust}, against
our proposal tracking strategy, PLVO-L1, when employing LSD or FLD features.
\begin{table}[tb]
\centering
\caption[]{Relative RMSE errors in the EuRoC MAV dataset \cite{burri2016euroc}.}
\label{tab_eurocodo}
\resizebox{\columnwidth}{!}{%
\begin{tabular}{l|cccc}
Sequence          & {LVO (FLD)} & {LVO-L1 (FLD)} & {LVO (LSD)} & {LVO-L1 (LSD)} \\ 
\hline
\hline \Tstrut
MH-01-easy  	  & 0.0641 & 0.0788 & 0.0669 & 0.0716 \\
MH-02-easy  	  & 0.0826 & 0.0923 & 0.0740 & 0.0881 \\
MH-03-med	      & 0.0886 & 0.1011 & 0.0898 & 0.1004 \\
MH-04-diff        & 0.1500 & 0.1536 & 0.1429 & 0.1518 \\
MH-05-diff        & 0.1350 & 0.1529 & 0.1391 & 0.1561 \\
V1-01-easy  	  & 0.0890 & 0.0969 & 0.0876 & 0.0954 \\
V1-02-med  	      & 0.0662 & 0.0847 & 0.0606 & 0.0947 \\
V1-03-diff	      & 0.2261 & 0.1518 & 0.0765 & 0.1103 \\
V2-01-easy  	  & 0.1980 & 0.1868 & 0.1662 & 0.1898 \\
V2-02-med  	      & 0.1634 & 0.2294 & 0.1982 & 0.2562 \\
V2-03-diff	      & 0.2329 & 0.2342 & 0.2354 & 0.2275 \\
\hline \Tstrut
MH-01-easy*  	  & 0.0787 & 0.0897 & 0.0741 & 0.0728 \\
MH-02-easy*  	  & 0.0873 & 0.1015 & 0.8237 & 0.0981 \\
MH-03-med*	      & 0.0982 & 0.1578 & 0.0916 & 0.1141 \\
MH-04-diff*       & 0.1540 & 0.1780 & 0.1354 & 0.1621 \\
MH-05-diff*       & -      & 0.1603 & -      & 0.1863 \\
V1-01-easy*  	  & 0.0880 & 0.1011 & 0.0997 & 0.1041 \\
V1-02-med* 	      & 0.0858 & 0.0953 & 0.0713 & 0.1096 \\
V1-03-diff*	      & -      & 0.2087 & -      & 0.1598 \\
V2-01-easy*  	  & -      & 0.2396 & -      & 0.2080 \\
V2-02-med* 	      & -      & 0.2472 & -      & 0.2563 \\
V2-03-diff*	      & -      & -      & -      & 0.2631            
\end{tabular}
\vspace{\indtab}
}
\end{table}

\tab{tab_eurocodo} contains the results by computing the relative RMSE in translation for the
estimated trajectories.
As we can observe, in the raw dataset our approach performs slightly worse than standard appearance-based 
tracking techniques, 
mainly due to the lower number of correspondences provided by our algorithm, as mentioned in previous Section.
In contrast, we can observe a considerable decrease in accuracy of our approaches, however,
they are capable of estimating the motion in all sequences with an lower accuracy, mainly due to 
the less number of matches, due to restrictive constraints.
For this reason we believe our matching technique a suitable option to address the line segment
tracking problem under severe appearance changes, in combination with prior information from
different sensors and/or algorithms.
%
%
%

\subsection{Computational Cost}
Finally, we compare the computational performance of the different tracking algorithms in the considered datasets considering
the time of processing one image (similarly to the VO framework).
In the both cases we can observe the superior performance of our proposal, it runs between 1.5 and 2 times
faster depending on the detector employed, thanks to the efficient implementation of the geometric-based tracking
thus making it very suitable for robust real-time application, 
most likely in combination with other sensing, such as inertial measurement unit sensors (IMU).

\section{{Conclusions}}
\label{sec_conclusions}
\begin{table}[tb]
\centering
\caption[]{Comparison of the computational performance of the different considered algorithms.}
\label{tab_perform}
\resizebox{\columnwidth}{!}
{
\begin{tabular}{lrr}
 &  \multicolumn{1}{c}{Monocular Tracking} &   \multicolumn{1}{c}{Stereo Tracking}      \\
\hline
\hline \Tstrut
LSD + LBD			     & 39.342 ms & 51.347 ms \\
LSD + Our           		 & 25.897 ms &  35.828 ms \\
\Tstrut
FLD + LBD			     & 18.147 ms & 33.266 ms \\
FLD + Our			     &  7.654 ms & 23.445 ms \\
\end{tabular}
}
\vspace{\indtab}
\end{table}

In this work, we have proposed a geometrical approach for the robust matching of line segments
for challenging stereo streams, such as sequences including severe illumination changes or 
HDR environments.
For that, we exploit the nature of the matching problem, i.e. every observation can  
only be corresponded with a single feature in the second image or not corresponded at all,
and hence we state the problem as a sparse, convex,  $\ell_{1}$-minimization
of the matching vector regularized by the geometric constraints.
Thanks to this formulation we are able of robustly tracking line segments along sequences 
recorded under dynamic changes in illumination conditions or in HDR scenarios where usual 
appearance-based matching techniques fail.
We validate the claimed features by first evaluating the matching performance in challenging
video sequences, and then testing the system in a benchmarked point and line based VO algorithm 
showing promising results.



\bibliographystyle{ieeetr}	
\bibliography{./biblio_cnn}

\end{document}